\documentclass[letter, 10 pt, conference]{ieeeconf}

\IEEEoverridecommandlockouts                              

\usepackage{xcolor}
\usepackage{graphicx}
\usepackage{xspace}
\usepackage{listings}
\usepackage{amsmath}
\usepackage{amsfonts}
\usepackage{tabularx}
\usepackage{caption}
\usepackage{subcaption}
\usepackage{cite}
\usepackage{hyperref}
\usepackage[capitalize]{cleveref}

\usepackage{mathrsfs}
\usepackage{bbm}
\usepackage{algorithm2e}
\usepackage{booktabs}
\usepackage{float}

\title{\LARGE \bf
Conveying Autonomous Robot Capabilities \\ through Contrasting Behaviour Summaries
}

\author{Peter Du, Surya Murthy, and Katherine Driggs-Campbell
\thanks{P. Du, S. Murthy, and K. Driggs-Campbell are with the Department of Electrical and Computer Engineering at the University of Illinois Urbana-Champaign. email: \{peterdu2,suryakm2,krdc\}@illinois.edu}%
}

\begin{document}

\maketitle


\begin{abstract}

As advances in artificial intelligence enable increasingly capable learning-based autonomous agents, it becomes more challenging for human observers to efficiently construct a mental model of the agent's behaviour. In order to successfully deploy autonomous agents, humans should not only be able to understand the individual limitations of the agents but also have insight on how they compare against one another. To do so, we need effective methods for generating human interpretable agent behaviour summaries. Single agent behaviour summarization has been tackled in the past through methods that generate explanations for why an agent chose to pick a particular action at a single timestep. However, for complex tasks, a per-action explanation may not be able to convey an agents global strategy. As a result, researchers have looked towards multi-timestep summaries which can better help humans assess an agents overall capability. More recently, multi-step summaries have also been used for generating contrasting examples to evaluate multiple agents. However, past approaches have largely relied on unstructured search methods to generate summaries and require agents to have a discrete action space. In this paper we present an adaptive search method for efficiently generating contrasting behaviour summaries with support for continuous state and action spaces. We perform a user study to evaluate the effectiveness of the summaries for helping humans discern the superior autonomous agent for a given task. Our results indicate that adaptive search can efficiently identify informative contrasting scenarios that enable humans to accurately select the better performing agent with a limited observation time budget. 

\end{abstract}


\section{Introduction}
\label{sec:introduction}

Autonomous agents are rapidly being developed and deployed in previously human dominated fields. Recent advancements in learning-based techniques has enabled autonomous agents to become increasingly more capable of performing complex tasks which may require interaction with other humans \cite{chen2019CrowdAwareNav, liu2021DSRNN, chen2017SociallyAwareMotionPlan, chen2017CADRL}. Today, these agents can be seen in diverse fields ranging from autonomous vehicles that drive among humans to robots that navigate through dense crowds. In order for system designers to deploy the appropriate agents that enable successful and safe human-autonomy interactions, we need methods for humans to understand the behaviour of the autonomous agents and identify any limitations in a particular agent's capabilities when compared with others. 

In the past, a number of works have been proposed to address the problem of communicating autonomous agent behaviour to humans. These works can be broadly classified into two categories: local and global. Local approaches look at a single state and aim to explain why an agent made a particular decision in that state. Local explanations for agent behaviour may fail to capture an agents overall strategy as certain behaviours may require a number of interdependent decisions spanning multiple states. Global approaches attempt to address this and instead explain an agents behaviour as a whole by communicating entire sequences of actions to human observers \cite{amir2018highlights, hayes2017ImproveRobotTransparency}. 

Given examples of an autonomous agent performing its task, humans will naturally start to familiarize themselves with the behaviour and improve their mental model of how the agent will act \cite{dragan2014FamiliarizationRobotMotion}. A straightforward but inefficient way of informing humans of an agent's global behaviour is to show them examples of the agent acting in various scenarios. However, this is not practical as human observers have a limited time budget to construct their mental model of the agent and not all scenarios are equally informative to the observer. For example, a scenario showing an autonomous vehicle driving down a straight road with no interactions with other drivers tells the human observer little about how the vehicle would react in a safety critical situation.

To address this, algorithms for generating more informative behaviour summaries based on state importance were proposed \cite{amir2019summarizingAgentStrats, amir2018highlights}. Modifications to these algorithms were made to tackle the agent comparison task by producing contrasting summaries involving multiple agents through the use of a disagreement metric \cite{amitai2022disagreements, gajcin2022contrastiveExplanations}. However, these approaches lack a more structured search method for exploring the scenario space (relying instead on randomization of the simulator or running in parallel during training of the agent) and do not generalize for agents with continuous actions.

In this paper we attend to these limitations and propose a structured search approach to contrastive behaviour summarization. We present the following contributions.

\begin{enumerate}
    \item We consider a global approach to behaviour summarization for helping human observers compare the capabilities of multiple autonomous agents. We relax the requirements of existing approaches which limit analysis to agents operating in a discrete action space. 
    
    \item We present an adaptive search based algorithm for generating contrastive behaviour summaries that can efficiently explore large, multi-agent scenario spaces.

    \item We use our methods to generate contrastive summaries for multiple learning-based crowd navigation robots. We perform a user study to evaluate if the summaries can help real humans efficiently and accurately discern differences in agent performance.
\end{enumerate}


\section{Background}
\label{sec:background}

\subsection{Explaining Agent Actions}

As advances in learning-based techniques have enabled autonomous agents to improve their performance in complex tasks, methods for explaining the decisions of these agents have also gained interest. Various works have analyzed ways for communicating pertinent behaviour characteristics of autonomous agents to humans with the goal of enabling humans to better understand the capabilities and limitations of the agent \cite{sequeira2020explainableRL, huang2019enablingRobotstoCommunicate, greydanus2018visualizeAtari}.

A number of ``local'' methods for agent behaviour explanation have been proposed \cite{khan2009minimalExplanationsMDP, khan2011automaticExplanationsMDP, elizalde2009explanationsMDP}. These methods typically target and try to explain an individual action the agent performs in a single state. While informative in certain scenarios such as critical states \cite{huang2018criticalStates}, local explanations may fail to effectively convey an agents overall strategy. 

When shown example sequences of an agent's behaviour, humans are able to learn the characteristics of the agent and form a mental model of how it will act \cite{dragan2014FamiliarizationRobotMotion}. Thus, there has been growing interest in ``global'' methods of behaviour summarization which aim to convey an agents global strategy to human observers rather than explain individual actions. Instead of providing a reason for why the agent chose a particular action, global methods seek to generate continuous sequences of actions (trajectories) that are informative for humans trying to construct a mental model. In \cite{amir2018highlights, amir2019summarizingAgentStrats}, the authors propose an algorithm for generating global behaviour summaries for single agents. The method relies on a state importance metric which measures the spread between an agent's Q-values at a particular state. During agent execution, states with high importance are collected to form candidate trajectories to show human observers.

While single agent summarization methods have shown to enhance the ability of humans to accurately discern the capabilities of agents, they are less effective at conveying which is superior when one wants to \textit{compare} among multiple agents. In these works, the summary for each agent is generated independently and may fail to capture meaningful scenarios where the behaviour of different agents diverge. Modified algorithms for generating behaviour summaries of two agents catered towards comparisons is presented in \cite{amitai2022disagreements} and \cite{gajcin2022contrastiveExplanations}. Instead of generating trajectories for single policies using state importance, the authors propose a coupled approach whereby the behaviour summary is determined by the disagreement of the two agents under comparison. This coupling enables the resulting summaries to demonstrate scenarios where the two agents choose different actions. 

A shared characteristic of all prior methods for global behavior summarization discussed thus far is their reliance on a discrete agent action space in order to evaluate the relevant metrics that determine which trajectories (summaries) they generate. For example, in \cite{amitai2022disagreements, gajcin2022contrastiveExplanations} the notion of a ``disagreement state'' $s_D$ (a state where two agents $A_1$ and $A_2$ with policies $\pi_1$ and $\pi_2$ such that $\pi_1(s) \neq \pi_2(s)$) is used to determine the start of the contrasting behaviour summary. For policies mapping to a continuous action space, such a definition of disagreement state becomes intractable as discrepancies in the policies result in virtually all states being classified as a disagreement state. This can be mitigated by adding an error threshold such that one requires $|\pi_1(s) - \pi_2(s)| < \epsilon$, however, the selection of $\epsilon$ may greatly affect the summaries produced and must be manually tuned for each application. 

The summaries produced are also highly reliant on how the user chooses to step the simulation containing the agents as the algorithms provide little guidance on how to explore the scenario space. In this paper, we consider using a divergence based metric for generating contrasting behaviour summaries to help humans compare autonomous agents. However, different from the prior approaches, we perform an adaptive search over the scenario space to maximize the divergence metric. By optimizing the metric over entire trajectories, we eliminate the restriction to discrete agent action spaces and demonstrate that we can efficiently generate informative contrasting behaviour summaries for agents with both a continuous state and action space. 

\subsection{Adaptive Scenario Search}

Adaptive search techniques have been applied to address the validation problem of black or semi-black box learning-based systems. Adaptive stress testing (AST) is one such technique which poses the problem of validation as an MDP and uses RL to generate a policy to produce relevant failure examples. Such a method has seen success when applied to a number of domains such as aircraft collision avoidance and autonomous vehicles \cite{koren2018adaptive}, \cite{lee2015adaptive}. AST is comprised of three component blocks consisting of a simulator $\mathscr{S}$, a reward function $R$, and RL solver $S$. The simulator contains the environment and the agent under test. The RL solver selects actions from a set of environment actions $\mathcal{A}$ to step the simulator. At each step, a reward is returned to the solver to update its policy. An example of the reward function used for adaptive failure search is:

\begin{equation}
\nonumber
\label{eq:ast_reward}
R = \left\{
        \begin{array}{ll}
            0 &  s \in E \\[7pt]
            -\alpha - \beta f(s) &  s \notin E, t\geq T \\[7pt]
            -g(a) - \eta h(s) &  s \notin E, t < T
        \end{array}
    \right.
\end{equation}

\noindent where $s$ is the current state (or observation) of the simulator, $E$ is a set of goal states for the search, $T$ is a finite time horizon, $g$ is an environment action based reward, $f$, $h$ are heuristic based rewards to guide the search process, and $\alpha$, $\beta$ are tuneable parameters. 

Subsequent works have explored multi-agent validation and reward augmentations to generate more informative failure modes. Differential AST compares two agents and searches for scenarios where one agent fails \cite{lee2018differential}. In this formulation, the standard single agent AST reward is used for the agent in which a failure is desired, while the second agent uses a negated version of the reward. As a result, the search tries to find scenarios where one agent ends in failure while the other agent does not. Reward augmentation and the use of critical states have also been applied to the AST formulation to improve the diversity and relevancy of the failure modes returned by the search \cite{corso2019ASTRewardAug} \cite{du2021ASTCriticalStates}. The prior works have shown that adaptive search can efficiently search large multi-agent state spaces and produce informative scenarios that convey the failure space of autonomous agents. In this paper, we take advantage of these proprieties of adaptive search and propose a method for generating contrasting behaviour summaries which allow human observers to accurately judge an agent's ability relative to others. 

\section{Methods}
\label{sec:methods}
\subsection{Coupled Simulation Environment}
Given two agents for which we wish to generate contrasting behaviour summaries, we create a coupled simulator $\mathscr{S}_c$ to perform adaptive search. Within the simulator, there are two separate simulation instances $\mathscr{S}_c^{(1)}$ and $\mathscr{S}_c^{(2)}$ for agents $\mathscr{A}^{(1)}$ and $\mathscr{A}^{(2)}$ with policies $\pi^{(1)}$ and $\pi^{(2)}$, respectively. The simulation instances contain all of the environment information required for the agents, as well as any environment actors that may interact with the agents. We represent the state of the simulations as follows: let $s_{env}$ denote the state of the simulation environment, and $s^{(i)}$ denote the state of the i-th agent. The coupled simulator $\mathscr{S}_c$ can be stepped forward in time by an environment action $a_{env}$. To ensure the behaviour summaries show examples of agents operating in the \textit{same} scenarios, $a_{env}$ is shared between both simulation instances $\mathscr{S}_c^{(1)}$ and $\mathscr{S}_c^{(2)}$. The environment action may alter any parameters or actions in the simulator, but not alter the agents themselves. Given a fixed sequence of environment actions, the evolution of the simulation instances are deterministic, allowing the two agents to experience the same environment. 

Propagating the simulation forward in time is accomplished in two steps. During the search process, the RL solver first generates an action $a_{env}$ from a fixed environment action space. The action is passed to both simulation instances within the simulator and is used to update the environment state $s_{env}$. Each agent then updates their own state $s^{(i)}$ according to their policy $\pi^{(i)}$. 

\subsection{Divergence-based Reward}
We design our search process to seek scenarios where the agents under comparison demonstrate substantially different behaviour despite acting in the same environment. The resulting scenarios are thus contrasting in nature. Work from the social sciences have argued that contrasting examples may be easier for humans to explain as one only needs to understand the \textit{differences} between two cases rather than determine all the causes for the scenario shown. Put another way, a layperson may find contrasting examples more intuitive and informative \cite{miller2019AISocialScience}. 

We consider two facets that contribute to the level of contrast between two agents in a given behaviour summary: state and action. At each step of the simulation, the divergence in the selected actions between agents will directly influence the perceived differences between their behaviour in the next timestep. Over the course of a trajectory, this biases the state of the agents to diverge and produce a high contrast summary. Here, we quantify the divergence of agent actions using the norm of the difference between action vectors. For any two agents ($\mathscr{A}^{(1)}$ and $\mathscr{A}^{(2)}$) under comparison, let $\pi^{(1)}$ and $\pi^{(2)}$ be the policies of the agents, respectively. Let the actions of the agents at a given environment state $s_{env}$ be generated such that $\pi^{(1)}(s_{env}), \pi^{(2)}(s_{env}) \in \mathbb{R}^{m}$. The action divergence $d_a$ at environment state $s_{env}$ is given by:

$$
d_{a}(s_{env}) = \sum_{i=1}^m{\Big[ \pi^{(1)}(s_{env})[i] - \pi^{(2)}(s_{env})[i] \Big] ^2}
$$

High action divergence alone is not sufficient for guiding the scenario search towards summaries with high contrast between agents. For example, consider a trajectory where two agents take the opposite action of each other at each timestep by alternating between actions $a$ and $-a$. The alternation between the positive and negative version of the action results in high action divergence, but the resulting states will have little contrast. When viewing the summaries, humans perceive highly contrastive behaviour through the states of the agents. Therefore, we choose the scenario search reward to take into account the state divergence $d_s$ at the end of the trajectory. Let the final agent states of the summary be $s^{(1)}, s^{(2)} \in \mathbb{R}^n$. The state divergence $d_s$ is given by:

$$
d_s(s^{(1)}, s^{(2)}) = \sum_{i=1}^n{ \Big[ s^{(1)}[i] - s^{(2)}[i] \Big] ^2}
$$

Combining the action and state divergences, the reward function used for adaptive search of contrasting behaviour summaries is as follows:

\begin{equation}
\nonumber
\label{eq:coupled_search_reward}
R = \left\{
        \begin{array}{ll}
            \alpha \cdot d_s(s^{(1)}, s^{(2)}) &  s^{(1)} \text{ or } s^{(2)} \in E \\[7pt]
            \beta \cdot h(s_{env}, s^{(1)}, s^{(2)}) &  s^{(1)}, s^{(2)} \notin E, t\geq T \\[7pt]
            d_a(s_{env}) &  s^{(1)}, s^{(2)} \notin E, t < T
        \end{array}
    \right.
\end{equation}

\noindent where $E$ is a set of agent goal states, $T$ is a finite search horizon, $h$ is an optional heuristic to guide search, and $\alpha$, $\beta$ are tunable constant parameters. 

\begin{figure}[!h]
    \centering
    \hspace*{-0.7cm}
    \includegraphics[width=0.725\columnwidth]{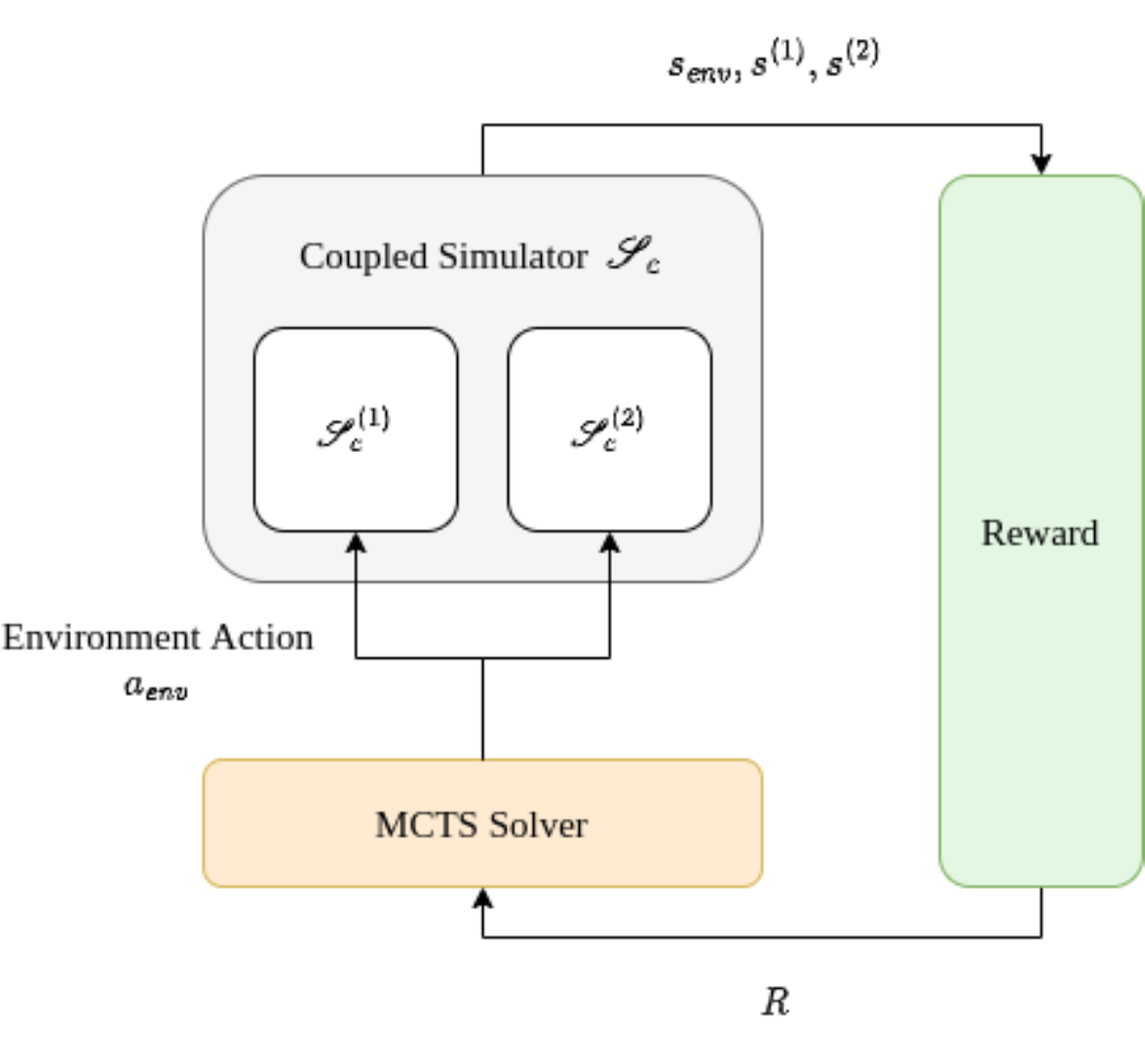}\\
    \caption{Adaptive Scenario Search}
    \label{fig:adaptive_scenario_search}
    \vspace{-0.2cm}
\end{figure}

\subsection{Adaptive Scenario Search}
Given $N$ agents $\mathscr{A}^{(1)}, ..., \mathscr{A}^{(N)}$, we generate contrasting behaviour summaries for all unique pairs of agents using $\frac{1}{2}N(N-1)$ instances of adaptive search. The block diagram of the search process is shown in \cref{fig:adaptive_scenario_search}. Before starting the search, we first create a priority queue $PQ$ of length $K$ which will contain trajectory sequences sorted by their corresponding cumulative reward. Next, we initialize an MCTS solver which will be used to generate environment actions to progress the simulation. At the beginning of each iteration of the search, the simulation environment and agents under comparison are reset to a fixed initial state $s_{init}$. During each timestep of an iteration, the MCTS solver steps the simulation forward, collects a reward signal from the reward function $R$, and updates its internal policy. If the simulation reaches a terminal state or the time horizon of the search is met, the current sequence of environment actions starting from $s_{init}$ are added to the priority queue $PQ$ and the next iteration of the search begins.

\RestyleAlgo{ruled}
\SetKwComment{Comment}{/* }{ */}

\begin{algorithm}[hbt!]
\label{alg:adaptive_scenario_search}
\caption{Adaptive Scenario Search}
\KwIn{$n_{itr}, s_{init}, \mathscr{S}_c, K, T$}
\KwOut{$X$}
$PQ \gets priority Queue(K)$\;
$S \gets solverMCTS$\;
$R \gets rewardFunction$\;
$i \gets 0$\;
\While{$i \leq n_{itr}$}{
    $step \gets 0$\;
    $T \gets emptyList$\;
    $\mathscr{S}_c.reset(s_{init})$\;
    \While{not $\mathscr{S}_c.isTerminal$ and $step < T$}{
        $a_{env} \gets S.getAction(\mathscr{S}_c)$\;
        $\mathscr{S}_c.step(a_{env}, \pi^{(1)}, \pi^{(2)})$\;
        $\mathcal{R} \gets R(\mathscr{S}_c)$\;
        $S.updatePolicy(\mathcal{R})$\;
        $T.append((a_{env}, \mathcal{R}))$\;
        $step \gets step + 1$\;
    }
    $trajReward \gets T.getReward + R(\mathscr{S}_c)$\; 
    $PQ.add((T, trajReward))$\;
    \If{$PQ.size \geq K$}{
        $PQ.popMin$
    }
    $i \gets i + 1$\;
}

$X \gets PQ.getTrajectories(K)$\;
\Return $X$
\end{algorithm}

\section{Evaluation}
\label{sec:evaluation}

\renewcommand{\arraystretch}{1.2}
\begin{table*}
        \caption{Accuracy Rate of Participants Correctly Ranking Agents}
        \label{table:results_accuracy_table}
        \centering
        \begin{tabular}{l|ccc|ccc}
            \toprule
            \text{} & & Part One (Pairwise Rankings) & & & Part Two (Global Rankings) \\\cline{1-7}
            Generation Scheme & Correct & Incorrect & \textbf{Accuracy} & Correct & Incorrect & \textbf{Accuracy} \\
            \midrule
            N-First Trajectories & 65 & 55 & \textbf{54.17}\% & 12 & 28 & \textbf{30.00}\%\\ 
            Adaptive Search & 114 & 6 & \textbf{95.00}\% & 26 & 14 & \textbf{65.00}\%\\
            \bottomrule
        \end{tabular}
        \vspace{-0.4cm}
\end{table*}

\subsection{Decentralized Structural-RNN Crowd Navigation}
\begin{figure}[!h]
    \centering
    \hspace*{0.15cm}
    \includegraphics[width=0.7\columnwidth]{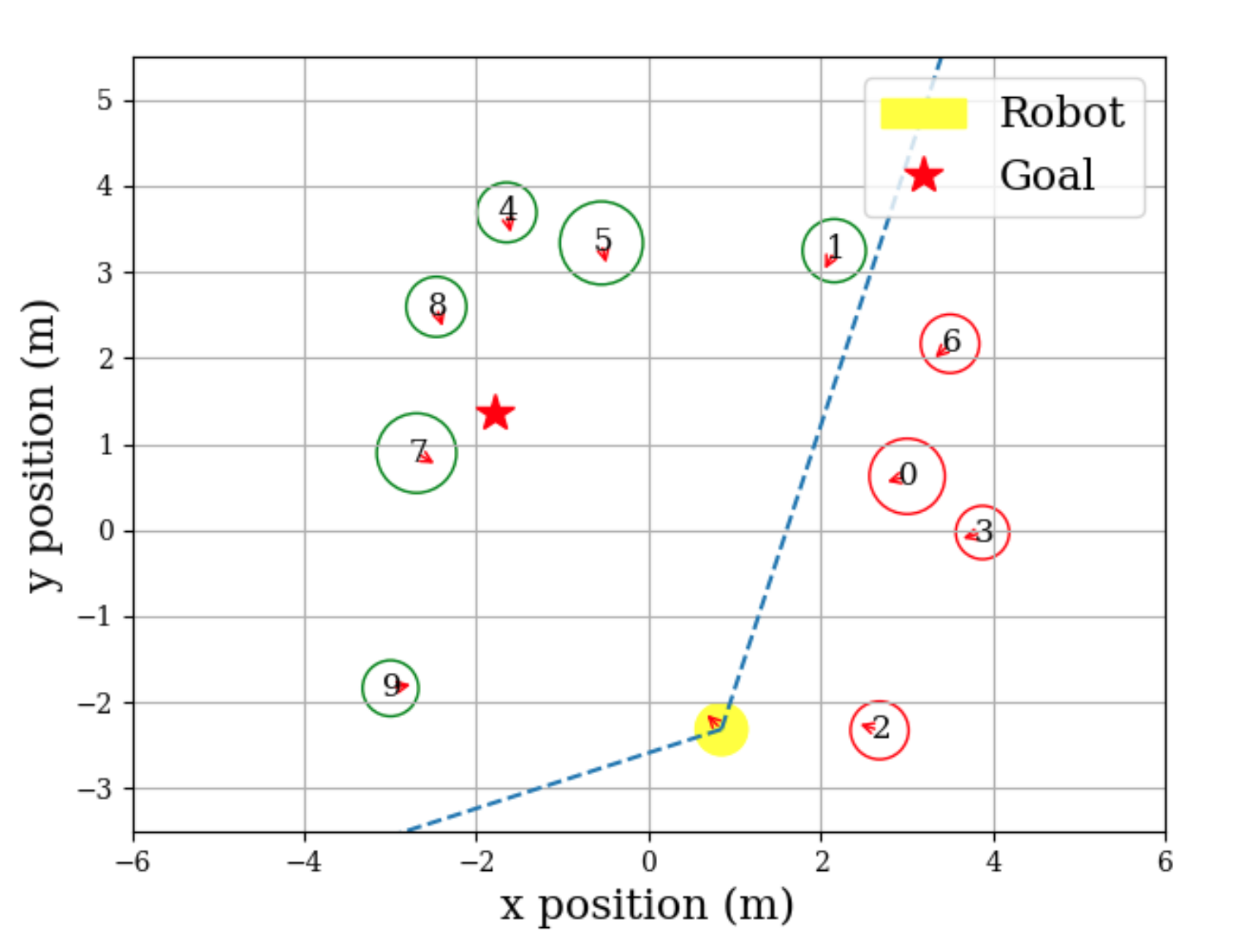}\\
    \caption{Robot Crowd Navigation Scenario}
    \label{fig:crowd_nav_scenario}
\end{figure}

We test the usage of adaptive search for contrasting behaviour summary generation on a robot crowd navigation scenario. An example of the scenario is shown in \cref{fig:crowd_nav_scenario}. In this setting, we have 10 environment agents (numbered 0 through 9) representing a human crowd. An autonomous robot is tasked with navigating to a goal state by going through the crowd and avoiding contact with the humans. 

We use a Decentralized Structural-Recurrent Neural Network (DS-RNN) RL agent which is trained to navigate to the goal state through dense crowds of humans \cite{liu2021DSRNN}. DS-RNN is a state-of-the-art method for robot crowd navigation which incorporates both spatial and temporal reasoning that allows the agent to safely and efficiently navigate through challenging situations in both simulation and the real world. 

In our evaluation, we train the DS-RNN crowd navigation agent using model-free deep reinforcement learning (DRL). We generate three agents to compare by selecting snapshots of the DS-RNN policy at three different stages in the training process, allowing us to have a set of agents with ground truth performance rankings. The agents are denoted as $\mathscr{A}^{(lo)}, \mathscr{A}^{(med)}, \mathscr{A}^{(hi)}$, corresponding to their performance from worst to best.

\renewcommand{\arraystretch}{1.2}
\begin{table}[H]
        \caption{DS-RNN Agents}
        \label{table:DSRNN_agents}
        \centering
        \begin{tabular}{l|c|c}
            \toprule
            Agent & Training Timesteps & Mean Episode Reward \\
            \midrule
            $\mathscr{A}^{(lo)}$ & $5.0 \times 10^6$ & $10.0$ \\
            $\mathscr{A}^{(med)}$ & $7.4 \times 10^6$ & $16.5$ \\
            $\mathscr{A}^{(hi)}$ & $1.2 \times 10^7$ & $21.5$ \\
            \bottomrule
        \end{tabular}
\end{table}

\subsection{Behaviour Summary Generation}
\label{subsec:behavior_summary_generation}
We generate contrasting behaviour summaries using adaptive search for all pairs of DS-RNN agents: $(\mathscr{A}^{(lo)}, \mathscr{A}^{(med)})$, $(\mathscr{A}^{(med)}, \mathscr{A}^{(hi)})$, $(\mathscr{A}^{(lo)}, \mathscr{A}^{(hi)})$. For each pair, we consider two initial states which represent different scenarios the agent may be asked to navigate in. The first scenario tasks the robot with navigating from the bottom left to the top right corner of the space, while the second has the robot go from the top left to bottom right corner. We instantiate an independent instance of adaptive search for each agent pair and initial state. At the end of the search, we select the highest reward trajectory from the priority queue to be the behaviour summary for that instance. As a result, we generate a set of summaries consisting of 6 pairwise contrasting trajectories. 

In addition to the set obtained from adaptive search, we generate a set of baseline (N-first) summaries by uniformly sampling the space of environment actions. This approach is similar to the case where a human observer is trying to build a mental model of the agents by watching a number of trajectories within a fixed viewing budget. As the trajectories are uniformly sampled, the observer will see agent behaviour that is more frequently exhibited. To match the output of adaptive search, we select a viewing budget of 6 summaries.

\subsection{User Study Procedure}
A user study is performed to evaluate the efficacy of the behaviour summaries. A total of 40 participants (50/50 Male/Female split) were recruited via Amazon Mechanical Turk and tasked with responding to questions in a survey format. Prior to beginning, participants were shown an example of the robot crowd navigation task to ensure understanding of the scenario. As stated in \cref{subsec:behavior_summary_generation}, we consider two sets of initial and goal states to represent two unique scenarios. At the start of the survey, the participants are asked to randomly select a value between 1 and 4 which is used to assign them to a particular scenario. In order to incentivize completion, participants were compensated with USD \$4 upon successful submission of the survey. 

The user study survey is comprised of two parts. In the first part, we evaluate the ability to discern the superior agent in pairwise examples given a contrasting behaviour summary. The participants are asked to view summaries in the form of short video clips, each containing a pair of robots navigating to the same goal. Based on the behaviour of the observed robotic agents, participants are asked to select the superior agent from the pair, as well as provide a brief justification for why they made that decision. Afterwards, the participants are asked to quantify their perceived difficulty in discerning the better agent using a 9-point semantic differential scale ranging from ``Very Easy" to ``Very Hard." This procedure is repeated for all 6 contrasting summaries. 

In the second part of the survey, we consider multi-agent rankings using pairwise contrasting behaviour summaries. The participants are shown three video clips covering all pairwise summaries of the three DS-RNN agents ($lo$, $med$, $hi$). After viewing the videos, participants are asked to rank the agents from best to worst based off the behaviours they observed. In addition, we request a short explanation for why they chose that particular ranking. The participants are also asked to rate the perceived differences between the agents using a 9-point semantic differential scale ranging from ``Very Small" to ``Very Large." Lastly, they are asked to rate their agreement to the statement that it was easy to distinguish between the agent pairs using a 5-point Likert scale. In all examples, the naming of the DS-RNN agents were given generic numerical values to avoid biasing the participants to any particular agent. The contrasting behaviour summaries shown in both part one and part two of the user study have unique initial and goal positions to maintain independence of the two parts during the survey.

\section{Results}
\label{sec:results}

In this section we present the results of our user study comparing the use of adaptive search versus passive viewing for agent behaviour comparison.

\begin{figure}[b]
    \centering
    \includegraphics[width=0.90\columnwidth]{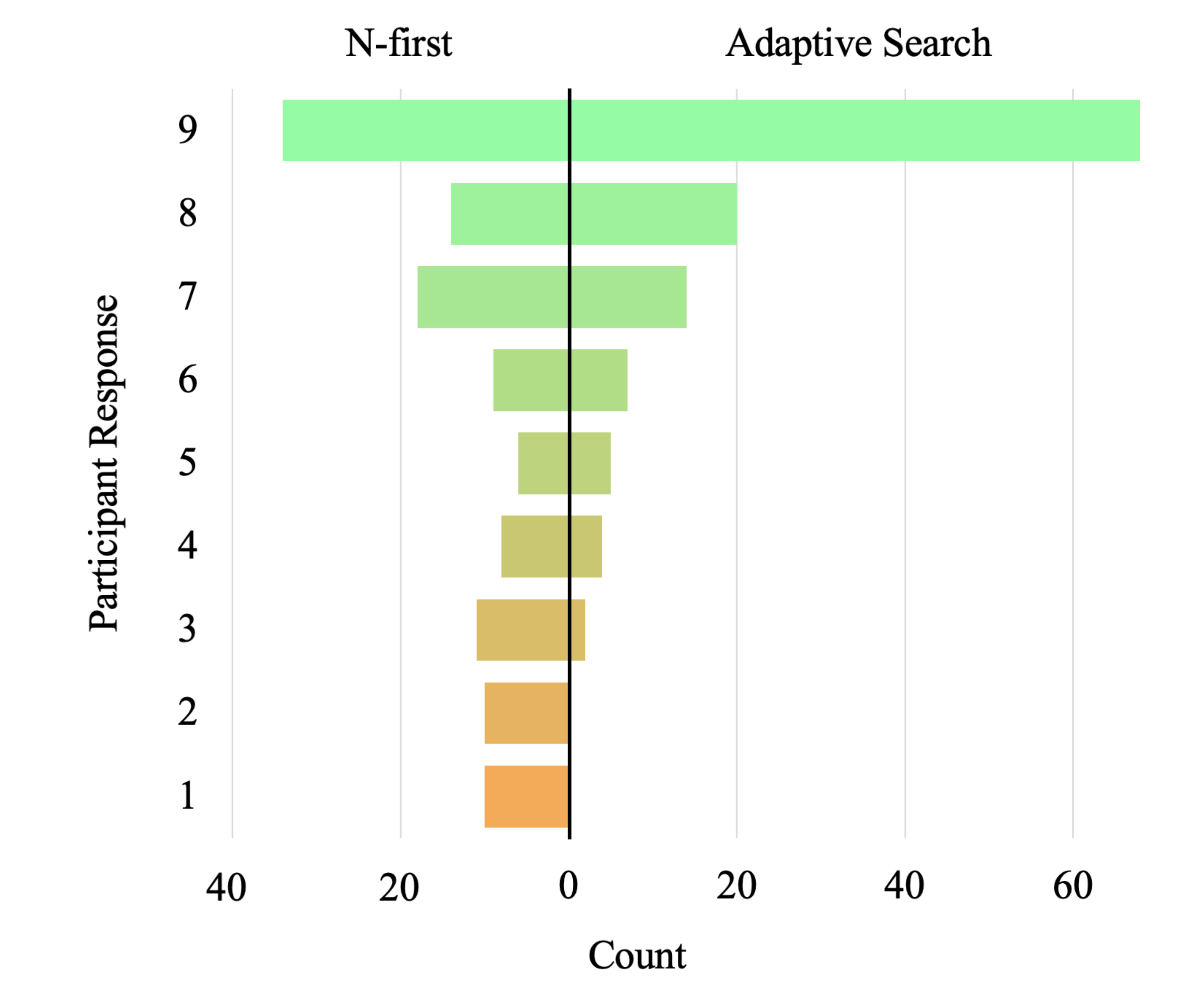}\\
    \caption{Participant responses to task difficulty in part one of survey (1 = Very Hard, 9 = Very Easy).}
    \label{fig:results_part_one_pref}
\end{figure}

\begin{figure*}[t]
    \centering
    \includegraphics[width=2.0\columnwidth]{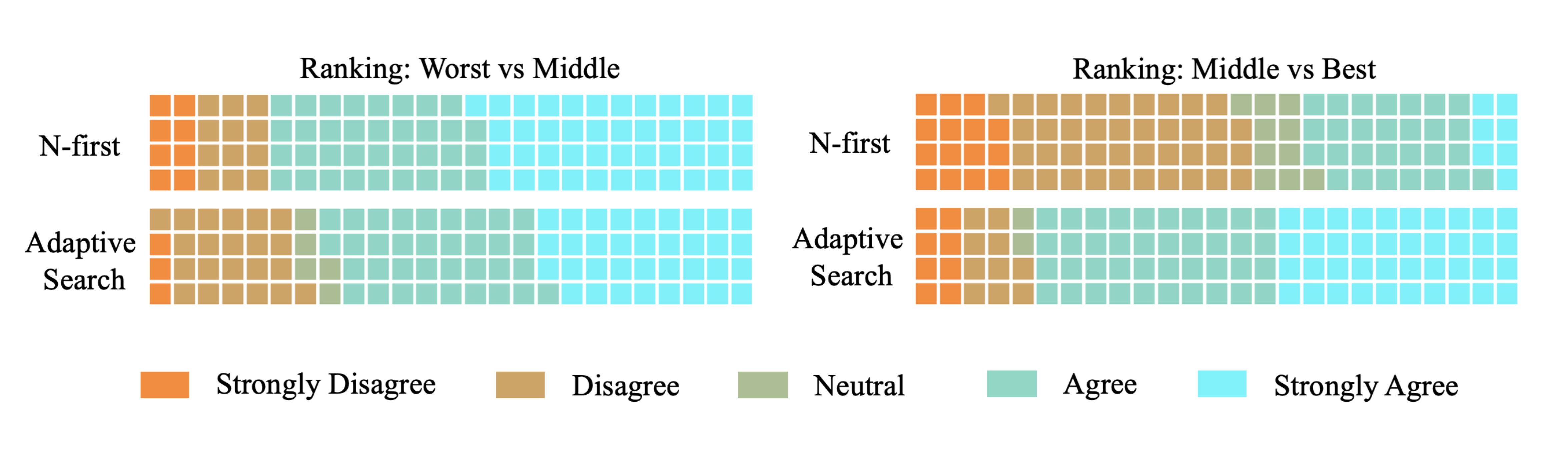}\\
    \caption{Participant agreement with the statement: It was easy to distinguish between the worst ranked and middle ranked robot (middle ranked and best ranked robot). Orange indicates more disagreement, blue indicates more agreement.}
    \label{fig:results_likert_waffle}
    \vspace{-0.2cm}
\end{figure*}

\cref{table:results_accuracy_table} shows the results of the users ability to correctly identify the superior policy and rank them in the correct order given contrasting behaviour summaries by the N-first and adaptive search generation schemes. We observe that the pairwise ranking task in part one of the survey is easier for participants. When given a single contrasting summary showing the behaviour of two robots, participants are able to correctly identify the superior agent with high accuracy when using the adaptive search generation scheme and moderate accuracy when using N-first. Part two of the survey presents a more difficult task by asking participants to rank three different robots given pairwise behaviour summaries. Here, a user response is marked as correct if they rank all three DS-RNN agents in the correct order. For this task, the accuracy of both generation schemes suffers when compared to part one. When using pairwise summaries given by the N-first scheme, participants achieve poor accuracy with less than a third of respondents able to correctly rank the robotic agents. However, in contrast, when using summaries generated by adaptive search, the users are still able to achieve moderate accuracy with approximately 2/3rds of respondents correctly ranking all agents.

\cref{fig:results_part_one_pref} plots the responses of perceived difficulty for part one (pairwise rankings) of the survey. After viewing each behaviour summary and selecting which agent they perceive as superior, the participants are asked to rate how difficult it was to discern the superior robot from the video summary. We see a clear advantage to perceived difficulty when using contrasting summaries generated by adaptive search. A majority of the participants (56\% of collected responses) responded that it was ``Very Easy" to discern the better robot given the summary they viewed, while no ``Very Hard" responses were logged. In comparison, only 28\% of the responses when using N-first characterized the difficulty as ``Very Easy" and 8\% stated it was ``Very Hard" to discern between the robots.



In part two of the survey the participants were asked for their agreement of whether it was easy to distinguish between each agent pair. In each global ranking, the user was asked to order the agents from best to middle to worst. As a result, each user was asked the above question for two pairs: the worst and middle ranked agents, and the middle and best ranked agents. \cref{fig:results_likert_waffle} shows the agreement responses for the global ranking task. We observe that when using summaries generated by N-first to select between the (perceived) worst and middle ranked agents, a majority of participants (80\%) agreed that the summaries offered enough contrast that it was easy to distinguish the better policy. However, the unstructured generation of summaries suffers when used to distinguish between the middle and best ranked policy with only 35\% of participants agreeing that it was easy to distinguish between the two robots. This indicates that beyond a certain performance threshold, simply observing the behaviour of agents may become less effective for determining their true capability as the agents all start to perform at a similar level for a large set of the scenario space. When using summaries generated by adaptive search, we see no perceptible improvement in the users abilities to discern between the worst and middle ranked policies when compared to N-first --- with approximately 70\% of those surveyed selecting agree or strongly agree. However, a noticeable discrepancy in participant responses between adaptive search and N-first is observed when looking at how difficult it was to discern the middle and best ranked robots. When using adaptive search, the behaviour summaries are able to continue to provide high contrast trajectories even when agent performance begins to converge resulting in approximately 80\% participant agreement. In contrast, we observe that a majority of participants disagreed that it was easy to discern between the two robots when using N-first summaries (indicated by the large orange region of the top right plot of \cref{fig:results_likert_waffle}). This result emphasizes the need for efficient exploration methods for generating informative contrasting behaviour summaries. As the differing policies begin to achieve high levels of performance, an unstructured search over the scenario space becomes less effective at uncovering high contrast scenarios of agent behaviour divergence.


\section{Conclusion}
\label{sec:conclusion}

As the development of various learning algorithms produce a variety of high performing autonomous agents, selection of the appropriate agent to deploy for a given task becomes increasingly more difficult. Passive or unstructured methods of observing the behaviour of autonomous agents become time-inefficient for determining how the performance of agents differ as the space of high divergence scenarios reduces as agent behaviour comes closer to optimality. In this paper, we presented an adaptive search approach for contrasting behaviour summary generation where we seek to uncover scenarios of high divergence between pairs of agents in a time efficient manner. Experimental results from our user studies indicate that summaries generated through such an approach can effectively convey to human observers the better robotic agent in an easily discernable manner. 

We enforced that the coupled simulation environment consider pairs of agents. This reduces the space of scenarios for each search instantiation at the expense of requiring multiple runs of the search process when comparing $N > 2$ agents. In future work we would like to expand the simulation environment to support all $N$ agents in one search instantiation. By coupling all agents, the resulting summaries may be able to more accurately convey global rankings. 




\bibliographystyle{IEEEtran}
\bibliography{references}

\end{document}